\documentclass{article} % For LaTeX2e
\usepackage{iclr2025_conference,times}

\usepackage{amsmath,amsfonts,bm}

\def\eqref#1{equation~\ref{#1}}
\def\1{\bm{1}}

\DeclareMathAlphabet{\mathsfit}{\encodingdefault}{\sfdefault}{m}{sl}
\SetMathAlphabet{\mathsfit}{bold}{\encodingdefault}{\sfdefault}{bx}{n}

\usepackage{soul}      % \ul 命令需要此包
\usepackage{siunitx}   % S 列需要此包
\usepackage{booktabs}  % \toprule, \midrule, \bottomrule 等命令需要此包
\usepackage{amsmath} % For mathematical expressions
\usepackage{algorithm} % For the algorithm environment
\usepackage{algpseudocode} % For the pseudocode environment
\usepackage{wrapfig}
\usepackage{tcolorbox}

\usepackage{amsthm}
\usepackage{amsmath}
\usepackage{algorithm}
\usepackage{xcolor}
\usepackage{makecell}  % \makecell 命令需要此包
\usepackage{adjustbox}
\usepackage{siunitx}
\usepackage{soul}
\usepackage{hyperref}
\usepackage{url}
\usepackage[detect-weight, detect-mode]{siunitx}
\usepackage[a4paper, margin=1in]{geometry} % 设置页面边距，确保表格有足够空间
\usepackage{booktabs}                     % 提供高质量的表格线 (toprule, midrule, bottomrule)                  % 用于颜色定义
\usepackage{colortbl}                     % 用于给单元格上色
\usepackage{graphicx}                     % 用于旋转等高级功能
\usepackage{makecell}                     % 用于在单元格内换行和精细控制
\usepackage{comment}
\usepackage{multirow}
\usepackage{amsmath}                      % 数学公式包
\usepackage{amsfonts}                     % 数学字体包
\usepackage{adjustbox}
\usepackage{booktabs}   % 用于三线表
\usepackage{makecell}   % 用于单元格内换行
\usepackage{siunitx}    % 用于S列类型，实现小数点对齐
\usepackage{caption}    % 改善标题

\newtheorem{proposition}{Proposition}
\usepackage{xcolor}
\definecolor{highlightgray}{gray}{0.9}
\definecolor{SectionGray}{gray}{0.93}
\definecolor{rowblue}{HTML}{E7F6FF}
\definecolor{rowgreen}{HTML}{E6F8F0}
\definecolor{rowgray}{HTML}{F3F4F6}
\definecolor{MyGray}{gray}{0.92}
\definecolor{MySkyBlue}{rgb}{0.9, 0.95, 1.0}

\setcellgapes{3pt}
\makegapedcells

\title{\name: Multi-level Stepwise Hints Enhance Reinforcement Learning to Reason}

\author{%
  Kaiyi Zhang$^{1,3}$, Ang Lv$^{1}$, Jinpeng Li$^{2}$, Yongbo Wang$^{3}$, Feng Wang$^{3}$, Haoyuan Hu$^{3}$ , Rui Yan$^{4}$ \\
  $^{1}$GSAI, Renmin University of China, 
  $^{2}$Peking University,
  $^{3}$Ant Group, 
  $^{4}$SCS, Wuhan University \\
  \texttt{{kyzhang@ruc.edu.cn}}, \quad \texttt{ruiyan@whu.edu.cn}\\
}

\newcommand{\name}{StepHint}

\iclrfinalcopy % Uncomment for camera-ready version, but NOT for submission.
\begin{document}

\maketitle

\begin{abstract}
Reinforcement learning with verifiable rewards (RLVR) is a promising approach for improving the complex reasoning abilities of large language models (LLMs). However, current RLVR methods face two significant challenges: the near-miss reward problem, where a small mistake can invalidate an otherwise correct reasoning process, greatly hindering training efficiency; and exploration stagnation, where models tend to focus on solutions within their ``comfort zone,'' lacking the motivation to explore potentially more effective alternatives.
To address these challenges, we propose \name, a novel RLVR algorithm that utilizes multi-level stepwise hints to help models explore the solution space more effectively. 
\name~generates valid reasoning chains from stronger models and partitions these chains into reasoning steps using our proposed adaptive partitioning method.
The initial few steps are used as hints, and simultaneously, multiple-level hints (each comprising a different number of steps) are provided to the model.
This approach directs the model's exploration toward a promising solution subspace while preserving its flexibility for independent exploration.
By providing hints, \name~mitigates the near-miss reward problem, thereby improving training efficiency. Additionally, the external reasoning pathways help the model develop better reasoning abilities, enabling it to move beyond its ``comfort zone'' and mitigate exploration stagnation.
\name\ outperforms competitive RLVR enhancement methods across six mathematical benchmarks, while also demonstrating superior generalization and excelling over baselines on out-of-domain benchmarks.
\end{abstract}
% \textcolor{red}{A key innovation is the multi-level structure of these hints, which is designed so that at least one hint level aligns with the policy model's current ability, striking the balance between effective guidance and independent exploration.

\section{Introduction}

Eliciting the reasoning capabilities of large language models (LLMs) through Reinforcement Learning with Verifiable Rewards (RLVR) has emerged as a powerful paradigm~\citep{jaech2024openai, guo2025deepseek}. In RLVR frameworks, a policy model explores the solution space by generating reasoning chains. The model is then optimized using algorithms like PPO~\citep{schulman2017proximal} and GRPO~\citep{shao2024deepseekmath}, based on the advantages of final outcomes of these chains.

However, free exploration within the vast and complex solution space introduces significant challenges. 
A key issue is the \textbf{near-miss reward problem}, where a single incorrect step voids an otherwise reward-worthy reasoning chain. 
This leads to training inefficiency, as models expend resources on repeatedly almost-correct solutions.
Moreover, as shown by~\citet{yue2025does}, existing RLVR methods often refine the model's ability to sample known reasoning chains rather than discover novel or higher-quality ones. 
Consequently, when a task exceeds the model's current capabilities, it tends to remain confined to its ``comfort zone,'' unable to independently advance beyond familiar solutions—an issue we term \textbf{exploration stagnation}.

We propose \name, a novel augmented RLVR algorithm that integrates multi-level stepwise hints to address these challenges.
\name\ leverages reasoning chains from advanced models such as Deepseek-R1~\citep{guo2025deepseek}, partitioning them into discrete reasoning steps \footnote{A \textit{reasoning step} refers to a distinct logical stage within the overall reasoning chain and typically comprises multiple tokens. 
It should not be confused with a \textit{token-prediction step} during generation or a \textit{training step}.}. 
It then provides only the initial few steps as hints for the model to complete the reasoning process. 
This approach effectively simplifies the solution space while preserving sufficient exploratory flexibility.
Specifically, during RLVR—regardless of the policy optimization algorithm used—\name's pipeline comprises two key steps:

\paragraph{Step 1: Adaptive Stepwise Partitioning of Reasoning Chains.}

Previous methods often partition reasoning chains using superficial markers such as ``First'' or ``Step 1,'' which fail to reflect the actual hierarchical structure of reasoning~\citep{guo2025segment}. 
To overcome this, we introduce a next-token-probabilistic partitioning strategy that adaptively splits the chain into meaningful steps.
Our method estimates the model's probability of concluding the whole reasoning chain at each token—specifically, the likelihood of generating a designated end-of-reasoning token (e.g., \texttt{</think>}, which may vary by model).
A token is identified as a candidate endpoint if its probability of concluding the reasoning chain exceeds the probability of concluding the reasoning chain at the next token. 
The intuition is that the model's confidence in generating an end-of-reasoning token will be higher at the natural conclusion of a reasoning step than at the beginning of the subsequent step. From this set of candidates, we then randomly sample a number of points to serve as step endpoints, subject to a constraint on the minimum distance between them. These endpoints partition the entire reasoning chain into a predetermined number of steps. This enables flexible identification of coherent reasoning steps.
The initial few steps are then provided as hints to guide the model's rollouts during RL training.

\paragraph{Step 2: Multi-Level Hints for Problem Solving.}
The effectiveness of a hint depends on the number of reasoning steps it reveals. 
An effective hint must strike a balance—offering enough guidance, while still preserving space for free exploration. 
Overly detailed hints can reduce RL to a form of supervised fine-tuning (SFT), thereby limiting the model’s opportunity for independent exploration. 
Such exploration is crucial for activating the reasoning capabilities encoded in pre-trained models~\citep{lv2025climbcarveswisdomdeeper}. 
Furthermore, SFT has been shown to lead to weaker generalization~\citep{chu2025sft,chen2025sft}, particularly in training large reasoning models compared to RLVR. To formalize this trade-off, we define a hint's ``level'' as the number of initial reasoning steps it provides. Under this definition, a ``high-level'' hint is one that contains many steps, making it highly detailed. Such a detailed, high-level hint can render the problem-solving task trivial for the model, diminishing training efficacy. Conversely, a ``low-level'' hint, which contains very few steps, can be insufficient to guide the model, leaving it vulnerable to the ``near-miss reward problem.''
Because determining the optimal hint level for a given model-problem pair is inherently difficult, \name\ generates multi-level hints for each problem. This approach ensures that, with sufficiently fine-grained step partitioning, at least one hint level is likely to be suitable for the model's current reasoning ability.

By adaptively providing multi-level hints derived from advanced models, \name\ effectively addresses both the near-miss reward problem and exploration stagnation. 
First, the model receives appropriate guidance to complete reasoning chains correctly, significantly reducing near-miss rewards and improving training efficiency—leading to faster convergence.
Second, exposure to high-quality hints steers the policy toward more sophisticated reasoning patterns, preventing stagnation during independent exploration. 
This not only enhances the model's ability to break through its ``comfort zone'' but also avoids the poor generalization typical of SFT-based methods.

We evaluate \name\ by training a series of LLMs on mathematical tasks and comparing their performance against strong RLVR-enhanced baselines. Results demonstrate \name's effectiveness on both in-domain (math) and out-of-domain tasks.

$\bullet$ \textbf{In-domain performance}: Across six math benchmarks, \name\ surpasses existing methods by an average accuracy of 3.16 percentage points. 
Notably, it achieves significant improvements in pass@$k$ performance—a rigorous measure of reasoning abilities~\citep{yue2025does}—on two challenging benchmarks, AIME24 and AIME25~\citep{li2024numinamath}, even at large $k$ values.

$\bullet$  \textbf{Out-of-domain generalization}: \name\ also achieves the highest results on out-of-domain, non-mathematical benchmarks such as ARC-C~\citep{clark2018think} and GPQA-D~\citep{rein2024gpqa}, highlighting its robust generalization beyond its training domain.

\section{Background: Reinforcement Learning with Verifiable Rewards}

\label{rlvr}

The advancement of reasoning in LLMs has significantly benefited from Reinforcement Learning (RL) techniques~\citep{hu2025open,guo2025deepseek}. These methods provide a framework for models to learn optimal reasoning chain through reward-based feedback.

Reinforcement Learning with Verifiable Rewards (RLVR) is a specialized RL paradigm highly effective for training LLMs on tasks where the correctness of an outcome can be objectively verified, such as mathematical problem-solving or code generation. In the RLVR framework, the learning process is typically driven by automated, often binary (correct/incorrect), reward signals, which facilitates scalable self-improvement~\citep{gao2023scaling}. 

\subsection{Policy optimization algorithms}
To implement RLVR, policy optimization algorithms are used to adjust the LLM's generation policy ($\pi_\theta$) based on the reward signals. Here, we discuss the evolution from the widely-used Proximal Policy Optimization (PPO) to a more streamlined and efficient alternative, Group Relative Policy Optimization (GRPO).
 
\paragraph{Proximal Policy Optimization (PPO)}
PPO~\citep{schulman2017proximal} is a popular policy optimization algorithm. 
PPO aims to maximize the expected reward while preventing excessively large policy updates that could destabilize the training process. 
Given a problem $q$, the policy model $\pi_{\theta}$ samples $N$ rollouts, denoted as $\{y_1,y_2,\cdots,y_N\}$. 
PPO optimizes the policy by maximizing the following objective function:

\[
\mathcal{L}_{\theta}^{\mathrm{PPO}} = \frac{1}{N}\sum_{i=1}^{N}\frac{1}{|y_i|}\sum_{t=1}^{|y_i|} \left\{
\min \left[ r_{i,t}\hat{A}_{i,t} ,\text{clip}\left( r_{i,t},1-\epsilon,1+\epsilon\right)\hat{A}_{i,t} \right]
\right\} - \beta D_{KL}(\pi_{\theta} || \pi_{\text{ref}}),
\]

where:
\begin{itemize}
\item $
r_{i,t} = \frac{\pi_{\theta}(y_{i,t}|q, y_{i,<t})}{\pi_{\text{old}}(y_{i,t}|q, y_{i,<t})}
$ is the probability ratio between the current policy $\pi_\theta$ and the policy before the update, $\pi_{old}$. The clip function bounds this ratio, which disincentivizes overly aggressive policy changes that could destabilize training.
\item $\hat{A}_{i,t}$ is the advantage of taking token $y_{i,t}$ as the $t$ token of rollout $i$. It quantifies how much better that action is compared to the average action value at that state.
\item $\beta D_{KL}(\pi_{\theta} || \pi_{\text{ref}})$ is a penalty term that discourages the current policy from deviating too far from a reference policy $\pi_{\text{ref}}$ (often the initial model).
\end{itemize}

In standard PPO implementation, calculating the advantage $\hat{A}_{i,t}$ for each token requires a critic model. This critic learns to estimate the value of states, often in conjunction with Generalized Advantage Estimation (GAE)~\citep{schulman2015high}, to compute the per-token advantages. 
However, PPO requires training a reliable critic model, which can be computationally expensive and complex.
This can lead to ``reward hacking,'' where the policy model learns to exploit the critic's inaccuracies to maximize rewards without achieving the actual task goal~\citep{amodei2016concrete}.

\paragraph{Group Relative Policy Optimization (GRPO)}
GRPO~\citep{shao2024deepseekmath}, a simpler yet effective alternative, was developed to address the challenges of PPO.
GRPO has gained significant traction for its remarkable performance and efficiency in complex reasoning tasks, especially mathematical reasoning tasks~\citep{liu2025understanding,yan2025learning,zeng2025simplerl,hu2025open}.

The core innovation of GRPO is to bypass the need for a critic model and per-token advantage calculation entirely. Instead, it computes a single, uniform advantage value for all tokens within a rollout, based on the final outcome of that entire rollout.

Specifically, for a given problem $q$, $N$ rollouts $\{y_1,y_2,\cdots,y_N\}$ are sampled. 
Each complete rollout $y_i$ is assigned a final reward $R(y_i)$, which is typically binary in RLVR settings: $R(y_i) = 1$ if the answer of $y_i$ is correct, and $R(y_i) = 0$ otherwise.
GRPO then calculates a rollout-level advantage by normalizing this reward across the group. This advantage value is assigned to every token within that rollout:
\begin{align}
\hat{A}_{i,t}^{\mathrm{GRPO}} = \frac{R(y_i) - \text{mean} \left( \left\{ R(y_1), \cdots, R(y_N) \right\} \right)}{\text{std} \left( \left\{ R(y_1), \cdots, R(y_N) \right\} \right)}.
\label{eq:adv-grpo}
\end{align}
This rollout-level advantage  $\hat{A}_{i,t}^{\mathrm{GRPO}}$ then replaces the complex, per-token advantage  $\hat{A}_{i,t}$ within the PPO objective function. The core clipping mechanism and KL-divergence penalty of PPO are retained, but the need for a separate critic model is eliminated entirely, drastically simplifying the training pipeline and reducing computational overhead.

The field of policy optimization includes other notable algorithms such as Trust Region Policy Optimization (TRPO)~\citep{schulman2015trust}, REINFORCE~\citep{sutton1999policy} and Asynchronous Advantage Actor-Critic (A3C)~\citep{mnih2016asynchronous}, PPO and GRPO have emerged as the most widely adopted methods for training LLMs on reasoning tasks. 
Therefore, the discussions and experiments in this paper will focus on these two algorithms as they represent the predominant approaches in this area.

\section{Method}

We conceptualize the reasoning process as a stepwise simplifying of a solution space. 
This perspective inspires our core idea: by providing stepwise hints, we can guide the model’s exploration toward more promising directions. 
To establish a foundation, Section~\ref{reasonchain} first formalizes the generation of reasoning chains in LLMs from a view of solution space exploration.
This formalization not only helps identify critical issues in existing methods, such as the near-miss reward problem and exploration stagnation, but also aids in better understanding our method, \name\ (Section~\ref{sec:method}), which is specifically designed to address these challenges.

\subsection{Preliminaries: A Solution Space View of Reasoning}

\label{reasonchain}

Autoregressive LLMs generate reasoning chains by producing tokens sequentially, where each new token depends on the preceding ones~\citep{radford2019language,grattafiori2024llama}.
This iterative simplification process is an exploration within a solution space $\mathcal{R}$. The space $\mathcal{R}$ consists of all possible reasoning chains the model might generate given a prompt $\mathcal{C}$~\citep{guo2025deepseek}. As tokens are generated, each token decision prunes the solution space, reducing the ambiguity and complexity of the final output.

To formalize this exploration process, we model it as a sequence of states. Each state, $S_k$, represents the partial reasoning chain constructed from the first $k$ generated tokens. The transition from one state to another corresponds to generating a new token. We quantify the complexity of the solution space at each state using conditional entropy $H(\mathcal{R}|S_k)$, which measures the remaining uncertainty given the tokens generated so far.

\begin{enumerate}

\item \textbf{Initial State ($S_0$):} 
The process begins with the prompt $\mathcal{C}$, which typically defines the problem to be solved in RLVR. 
This prompt constitutes the initial state, $S_0 = \mathcal{C}$. The initial complexity of the solution space, given only the prompt, is quantified by the conditional entropy $H(\mathcal{R}|\mathcal{C})$~\citep{shannon1948mathematical}. As the solution space is composed of discrete token sequences, this is computed as:
\[ H(\mathcal{R}|\mathcal{C}) = -\sum_{r\in\mathcal{R}} p(r|\mathcal{C})\log p(r|\mathcal{C}). \]
It is important to note that this formulation serves as a theoretical model for qualitative analysis. A direct quantitative computation of this entropy is generally intractable, as it would require summing over the entire space of all possible reasoning chains $\mathcal{R}$. Despite its intractability, this entropy-based framework provides a powerful conceptual tool for analysis.

A high value of $H(\mathcal{R}|\mathcal{C})$ indicates a complex and unconstrained solution space, while a low value suggests that the prompt has already constrained the problem to a narrower range.

\item \textbf{Intermediate States ($S_k$):} After generating $k$ tokens, $t_1, \dots, t_k$, the system transitions to an intermediate state $S_k = (\mathcal{C}, t_1, \dots, t_k)$. The remaining complexity of the solution space, given the generated tokens, is captured by the conditional entropy $H(\mathcal{R}|S_k) = H(\mathcal{R}|\mathcal{C}, t_1, \dots, t_k)$.

\end{enumerate}

The following Proposition formalizes that, in expectation, each token-prediction step reduces or maintains the entropy, which corresponds to the complexity of the solution space, often stated as ``conditioning reduces entropy.''~\citep{cover1999elements}

\begin{proposition}
\label{theo_chain}
Let $\mathcal{R}$ be the solution space and $S_{k-1}$ be the state after $k-1$ tokens have been generated. Upon generating the next token $t_k$ to form state $S_k = (S_{k-1}, t_k)$, the expected entropy of the solution space is bounded by the current entropy:
\[
\mathbb{E}_{t_{k} \sim P(\cdot|S_{k-1})}[H(\mathcal{R}|S_k)] \le H(\mathcal{R}|S_{k-1}).
\]
\end{proposition}

We leave the detailed proof in Appendix~\ref{proof1}. 
Proposition~\ref{theo_chain} provides a formal foundation for understanding autoregressive generation as a structured exploration. This process incrementally refines a vast solution space, converging on a specific output by reducing uncertainty and complexity at each step.
It should be noted that, the reduction in entropy quantifies the convergence of certainty, not necessarily the correctness or logical validity of the reasoning chain. A model can become increasingly certain about a flawed or nonsensical conclusion, and this would still manifest as a decrease in entropy. Therefore, entropy reduction should be understood as a necessary, but not sufficient, condition for reasoning. 
This view reveals a failure mode in reasoning, occurring when the model makes a critical early error. This misstep irrevocably prunes the subspace of correct solutions, $R^*$.
Formally, this corresponds to a state $S_k$ where the probability of the correct solution set collapses to zero: $P(R^*|S_k)$. 
The model may continue reducing entropy confidently, but within an incorrect subspace $R\backslash R^*$, leading to a ``confident but wrong'' conclusion.

\subsection{\name: Multi-level Stepwise Hints Enhance RLVR}
\label{sec:method}

Based on the views above, the reasoning efforts following the initial error might be logical and proper, but the answer remains wrong because they are based on an incorrect premise. 
This near-miss reward issue could be avoided if slight guidance or supervision were provided at the beginning. 
Meanwhile, the exploration of the solution space is largely constrained by the model's ability. 
Without external guidance, the exploration will be limited to a narrow subspace~\citep{yue2025does}. 
In this section, we introduce how our proposed RLVR enhancement method, \name, addresses these two issues by providing strong reasoning chains as hints to models during training.

Given a problem, \name\ first obtains a valid reasoning chain from a stronger model, and then performs two key stages to enhance the target model being trained: (1) adaptive stepwise partitioning of on-hand reasoning chains and (2) multi-level hints for problem solving.
The on-hand reasoning chains are collected offline by querying a stronger model before training the target model and retaining only outputs that have been verified as correct. 

In the following, we focus on detailing the latter two stages.
We will frequently use the term ``reasoning step'' to refer to an intact unit of thought, which may consist of several sentences. 
Please note that this is distinct from a ``training step,'' which refers to updating the model after processing a batch of data, or a ``next-token prediction step,'' which generates a single token at a time.

\subsubsection{Adaptive Stepwise Partitioning of On-Hand Reasoning Chains}
\label{sec:seg}

\paragraph{Definitions} 
Let the reasoning chain be denoted as $G = t_1\circ t_2\circ \dots \circ t_n$, where each $t_i$ represents a single token, and $\circ$ denotes concatenation. 
A \textit{reasoning step} corresponds to a sequence of tokens $t_i\circ \dots \circ t_j (1\leq i\leq j\leq n)$ that forms an intact unit of thought. 
A \textit{hint} is composed of one or more reasoning steps and serves as a conditioning prompt that guides the target model's reasoning toward a promising direction, helping it explore otherwise intractable solution spaces.
The \textit{level of a hint} is determined by the number of reasoning steps it contains—the more reasoning steps included, the richer the guidance it offers to the target model, and thus the higher its level.

\paragraph{Method Details} 
We need a flexible %and robust 
method to adaptively partition a complete reasoning chain $G$ into $m$ reasoning steps, then combine appropriate number of reasoning steps as a hint in appropriate level to the target model.
Conventional approach relies on syntactic cues, such as keywords like ``first,'' ``next,'' or ``Step 1.'' 
However, such heuristics are prone to misidentifying the boundaries of reasoning steps and lack the flexibility.~\citep{Moens2017ArgumentationMH}

In contrast, we leverage the model's output probability distribution to identify the boundaries of reasoning steps. 
We hypothesize that when a reasoning step concludes, the model's perceived probability of completing the entire chain should be relatively high. 
Conversely, at the beginning of a new step, this probability should decrease as the model expects additional reasoning to follow.
This perceived likelihood of reasoning completion can be captured by the probability assigned to a special ``end-of-thinking'' token, \texttt{</think>}, which is explicitly introduced during pretraining to mark the conclusion of a reasoning step~\citep{qwen2.5,guo2025deepseek}.
Formally, the model's tendency to conclude a reasoning step at token $t_i$ can be quantified by the probability $p(\texttt{</think>} \mid G_i)$, where $G_i$ denotes the token sequence up to $t_i$.

This hypothesis leads us to our core partitioning method: a token 
$t_i$ is considered as a candidate reasoning step boundary if and only if the probability of concluding the reasoning chain after $t_i$ is greater than the probability of 
\begin{wrapfigure}[18]{r}{0.65\textwidth}
\vspace{2mm}
  \centering
  \includegraphics[width=0.65\textwidth]{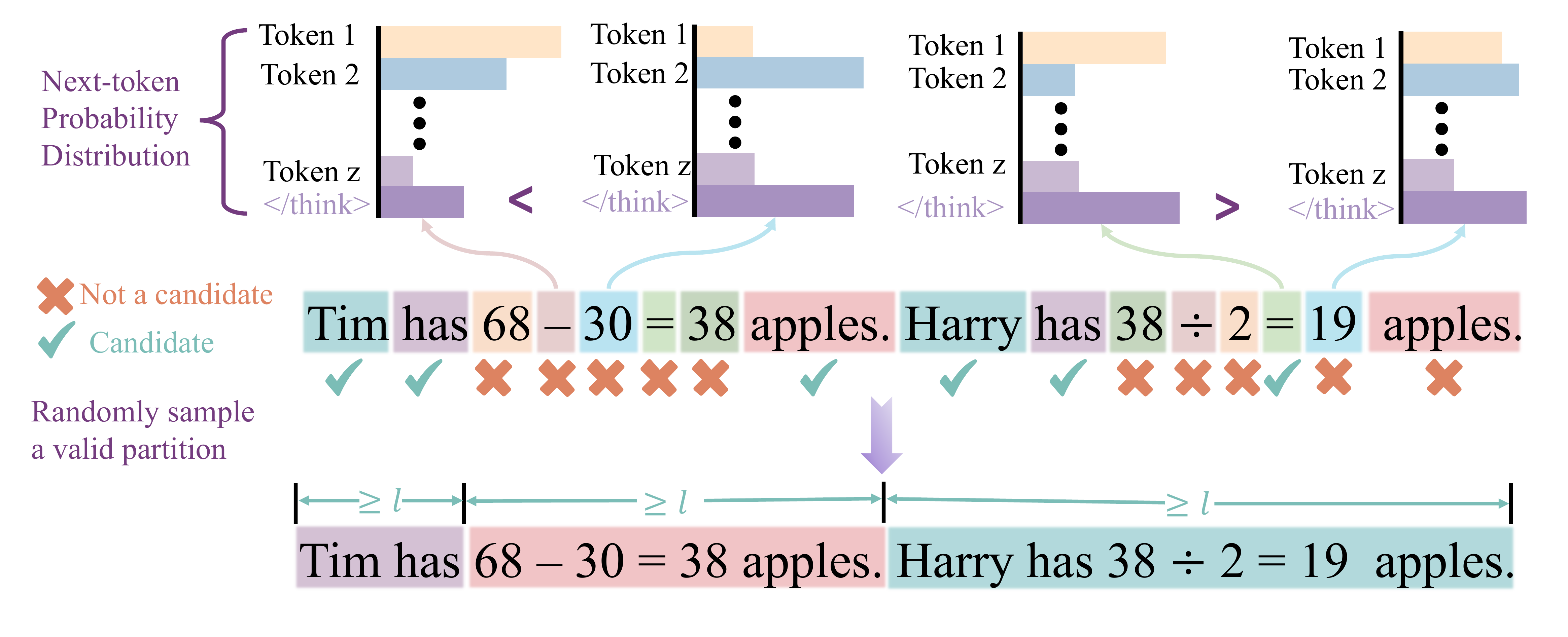}
  \caption{Adaptive stepwise partitioning of a reasoning chain: step boundaries are identified where the probability of concluding the reasoning chain after the current token, $p(\texttt{</think>}|G_i)$, is greater than concluding after the subsequent token, $p(\texttt{</think>}|G_{i+1})$. A final partition is chosen to meet constraints on step count $m$ and minimum length $l$.}
  \label{fig:par}
\end{wrapfigure}
concluding it after the subsequent token, $t_{i+1}$: $p(\texttt{</think>}|G_i) > p(\texttt{</think>}|G_{i+1})$.
By iterating through the entire reasoning chain, we collect all tokens satisfying this condition as candidate boundaries.
To maintain high-quality reasoning steps, we enforce two constraints during partitioning: (1) adjacent boundaries must be at least $l$ tokens apart to avoid overly short steps, and (2) 
the number of steps must be equal to the predetermined value, $m$.
In practice, multiple valid partitionings may satisfy these constraints, we randomly sample one to proceed with.
Figure~\ref{fig:par} illustrates this token-distribution-based partitioning method.

\subsubsection{Multi-level Hints for Problem Solving}
\label{sec:rollm}

\begin{figure}[t]
    \centering
    \includegraphics[width=1\textwidth]{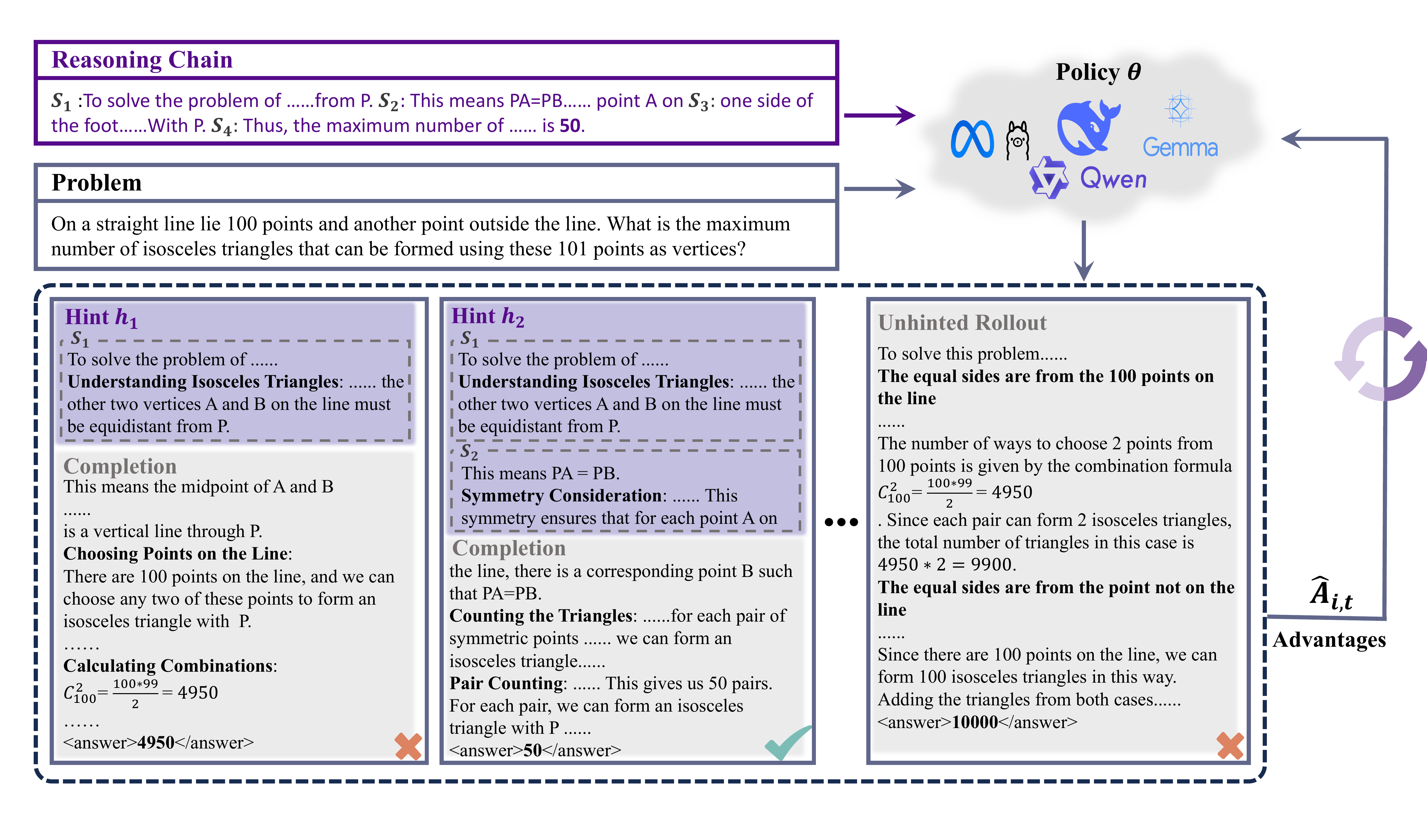} 
    \caption{
    An overview of the multi-level hinting process.
    The process begins with a ground-truth reasoning chain, which is partitioned into $m$ steps (Section~\ref{sec:seg}). 
    From these steps, we construct $m-1$ prefix-based hints$\left(h_1,h_2,\cdots,h_{m-1}\right)$. 
    The model is trained to generate completions from each hint level, as well as from scratch (Unhinted), and a reference trajectory. }
    \label{fig:spire}  
\end{figure}

Building on the adaptive stepwise partitioning method described above, we divide the reasoning chain into $m$ reasoning steps. 
A key question is how many of these steps should be included as a complete hint for the target model.

An ideal level of hinting matches the model's current capabilities at each stage of training—it is neither too weak nor so strong that it simplifies the task.
Moreover, this optimal level shifts continuously as the model's reasoning ability evolves.
Considering these challenges, rather than determining the ideal hint level at each training step, we provide hints at multiple levels.
Our core hypothesis is that if the partitioning is fine-grained enough, there is very likely to be a hint that fits the model's needs well.

Specifically, we construct a set of $m-1$ multi-level hints,  $\mathcal{H}$. 
Each hint is a prefix of the full reasoning chain $G$, created by concatenating the first $j$ steps:
\[
\mathcal{H} = \{h_j|h_j=S_1\circ S_2\circ \cdots\circ S_j,\quad \text{for}~j=1,\cdots,m-1\},
\]
where $S_i$ represents the $i$-th reasoning step. 
Low-level hints preserve considerable problem-solving difficulty.
In contrast, high-level hints significantly simplify the solution space, making the completion easier.

For each problem in the training set, we construct three types of prompts for the model to complete:

\begin{enumerate}
    \item \textbf{Hinted Problems:}
For each of the $m - 1$ hint levels, the model is asked to complete the reasoning from that hint using $k_{\text{hint}}$ rollout attempts per hint.
    
\item \textbf{Unhinted Problems:}  
To preserve the model's independent exploration, it also solves the problem from scratch without any hints. 
It is allowed \( k_{\text{unhint}} \) rollouts in this case.

    \item \textbf{Reference Trajectory:} 
    The original ground-truth reasoning chain $G$ is also provided to the target model and used to assign rewards.
This ensures that the model is consistently exposed to the correct solution path during training.
\end{enumerate}

Figure~\ref{fig:spire} illustrates this multi-level hinting and completion process.
In total, the model generates $k_{\text{hint}}(m-1) + k_{\text{unhint}} + 1$ completions per training problem, each receiving a reward based on correctness.
\name~strikes a balance between guiding the model with reliable hints and allowing it to learn from its own exploration mistakes, leading to more effective learning. 

The above method applies to most RLVR algorithms but poses issues for GRPO~\citep{he2025deepmath}, prompting further adaptations and discussion.
In GRPO, an incorrect completion assigns negative advantages to all tokens in the rollout—including those in the correct hint prefix. 
This steers the model away from the correct reasoning chains.
To address this, we modify GRPO by clipping negative advantages to zero for tokens in the hint prefix when the completion is incorrect; that is, we set $\hat{A}_{i,t}^{\mathrm{GRPO}} = \text{max}(0, \hat{A}_{i,t}^{\mathrm{GRPO}})$ (see Eq.~\ref{eq:adv-grpo} for the definition of $\hat{A}_{i,t}^{\mathrm{GRPO}}$). This adaptation prevents the model from being penalized for correct prefixes, aligning the training process with our intended mechanism. We demonstrate its empirical effectiveness in the next section.
\section{Experiments}
\subsection{Experimental Settings}
\paragraph{Training Data} 

To construct a scenario where the training problems are particularly challenging for the target model to independently explore, we gather difficult problems from the DAPO dataset~\citep{yu2025dapo} and selected problems of difficulty level 7 or higher from the DEEPMATH dataset~\citep{he2025deepmath}. 
The DEEPMATH dataset provides solution reasoning chains, while we generate reasoning chains for the DAPO dataset using the DAPO-Qwen-32B~\citep{yu2025dapo}, QWQ-32B~\citep{qwq32b}, and DeepSeek-R1-Distill-Qwen-32B~\citep{guo2025deepseek} models, as described in Appendix~\ref{dat}.
To ensure data quality, we filter out excessively long answers, retaining only those with a length of 4,096 tokens or fewer. 
All reasoning chains are partitioned into $m = 4$ steps, each longer than $l = \frac{L}{8}$ tokens, where $L$ is the total length of a reasoning chain $G$, with QWQ-32B~\citep{qwq32b}. This threshold ensures that even the shortest step contains a substantive amount of information, while allowing for the natural length variation between different steps in a complex reasoning process.

This results in a training dataset of approximately 26,000 instances.

\paragraph{Hyperparameters} 
We train two backbone models for 5 epochs each: Qwen-2.5-7B-Instruct~\citep{qwen2,qwen2.5} and Qwen-2.5-Math-7B~\citep{yang2024qwen25mathtechnicalreportmathematical}. 
The training prompt template is shown in Appendix~\ref{temp}.
The training is based on the VeRL framework~\citep{sheng2024hybridflow}. 
We set a global batch size of 128 and a fixed learning rate of $1e-6$. 
Following~\citep{yan2025learning}, we set the KL loss coefficient $\beta = 0$, indicating no reference model is used for regularization. 
We set $k_{\text{hint}} = 2$, and $k_{\text{unhint}} = 5$. 
During training, the temperature for rollout generation is set to 1.0.

\paragraph{Baselines} We compare against three categories of baselines:
\begin{enumerate}
\item \textbf{Vanilla GRPO:} The model is trained using vanilla GRPO on Qwen-2.5-7B-Instruct with the same settings as \name.
\item \textbf{SFT:} We also fine-tune the models with available reasoning chains using supervised fine-tuning (SFT), applying the same learning rate and training epochs as \name.
\item \textbf{Other RLVR Enhancement Methods:} We evaluate other RLVR enhancement techniques, including: SimpleRL-Zero-7B~\citep{zeng2025simplerl}, Qwen-2.5-Math-7B-SimpleRL-Zoo, OpenReasonerZero-7B~\citep{hu2025open}, Oat-7B~\citep{liu2025understanding}, Luffy-Qwen-2.5-7B-Instruct~\citep{yan2025learning}, and Luffy-Qwen-Math-7B-Zero. 
Evaluations are conducted using their publicly released model weights.
\end{enumerate}

\paragraph{Evaluation} We follow prior work~\citep{yan2025learning,liu2025understanding} and evaluate on six math datasets: AIME 2024, AIME 2025, AMC~\citep{li2024numinamath}, Minerva~\citep{lewkowycz2022solving}, OlympiadBench~\citep{he2024olympiadbench}, and MATH500~\citep{hendrycks2021measuring}. For the AIME 24/25 and AMC datasets, given their limited data points, we conduct each evaluation five times and report the average results. To evaluate generalization, we also incorporate two non-math benchmarks, ARC-C~\citep{clark2018think} and GPQA-Diamond~\citep{rein2024gpqa}, as out-of-domain tests for models trained on math problems. 
We report the weighted average accuracy for both in-domain and out-of-domain benchmarks. The generation length is also set to 4,110.
All results were evaluated using OAT-Grade~\citep{liu2024oat}.

\subsection{Main Results}

Table~\ref{tab:result_main} shows the overall performance of \name~and baseline methods.

When applied to the general-purpose model, Qwen-2.5-7B-Instruct, \name\ achieves the highest performance on in-domain math tasks among all compared methods. Compared to other RLVR methods, \name\ shows a 7 percentage point improvement over the next-best method, LUFFY. Furthermore, \name\ consistently surpasses the SFT baseline, indicating that \name\ effectively learns beyond simple token imitation, leading to improved reasoning outcomes. 
Notably, the Qwen-2.5-7B-Instruct model trained with \name\ outperforms the specialized Qwen-2.5-Math-7B fine-tuned with any other RLVR method, highlighting the substantial boost in reasoning ability provided by \name\ and allowing a generalist model to outperform a specialist in its own domain.

\begin{table}[t]
    \centering
    \caption{
    Performance comparison of \name~with baseline methods on in-domain and out-of-domain benchmarks.
    The top score in each column is in \textbf{bold}, and the second-highest is \underline{underlined}. Backbone models are denoted by: \textsuperscript{*}Qwen-2.5-7B-Instruct, \textsuperscript{†}Qwen-2.5-7B, \textsuperscript{‡}Qwen-2.5-Math-7B.
    }
    \label{tab:result_main}
    
    % The column specification is updated to remove the final "Total Avg." column.
    % @{\hskip 1em} adds a smaller horizontal space for better visual separation.
    \begin{adjustbox}{max width=\linewidth}
    \begin{tabular}{l *{6}{S} S @{\hskip 1em} *{2}{S} S}
        \toprule[1pt]
        
        % Super-headings for In-Domain and Out-of-Domain categories with their averages
        \multirow{4}{*}{\textbf{Model}}& \multicolumn{7}{c}{\textbf{In-Domain}} & \multicolumn{3}{c}{\textbf{Out-of-Domain}} \\
        \cmidrule(lr){2-8} \cmidrule(lr){9-11} % Lines under the super-headings
        
         & 
        {\makecell{\textbf{AIME24} \\ \scriptsize (avg@5)}} & 
        {\makecell{\textbf{MATH500} \\ \scriptsize (pass@1)}} & 
        {\makecell{\textbf{AMC} \\ \scriptsize (avg@5)}} & 
        {\makecell{\textbf{Olympiad} \\ \scriptsize (pass@1)}} & 
        {\makecell{\textbf{Minerva} \\ \scriptsize (pass@1)}} & 
        {\makecell{\textbf{AIME25} \\ \scriptsize (avg@5)}} &
        {\makecell{\textbf{Avg.} \\ \scriptsize -}} &
        {\makecell{\textbf{ARC-C} \\ \scriptsize (pass@1)}} & 
        {\makecell{\textbf{GPQA-D} \\ \scriptsize (pass@1)}} &
        {\makecell{\textbf{Avg.} \\ \scriptsize -}} \\
        
        \midrule % Rule between header and data
        
        SFT\textsuperscript{*}              & 20.00          & 78.80          & 53.73          & 36.89          & 36.40          & 10.67          & 53.76          & 90.96          & 23.23          & 83.28 \\
        SFT\textsuperscript{‡}              & 26.00          & 82.20          & 59.52          & 45.19          & 34.19          & 15.33          & 54.77          & 67.66          & 23.74          & 61.31 \\
        \midrule
        
        % The multicolumn span is decreased from 12 to 11 to cover the remaining columns
        \multicolumn{11}{l}{\textsf{\textbf{On-policy RLVR Replication}}} \\
        \addlinespace[2pt]
        Vanilla-GRPO\textsuperscript{*}       & 24.67          & 76.60          & 51.33          & 43.41          & \underline{39.34} & 10.67          & 52.59          & \underline{91.30} & 37.37          & \underline{83.51} \\
        
        \midrule
        
        \multicolumn{11}{l}{\textsf{\textbf{Other RLVR Methods}}} \\
        \addlinespace[2pt]
        ORZ-7B\textsuperscript{†}           & 24.67          & 81.00          & 46.90          & 45.60          & 33.46          & 15.30          & 53.76          & 90.53          & \underline{40.40} & 83.28 \\
        SimpleRL\textsuperscript{†}         & 22.00          & 76.00          & 47.90          & 39.30          & 36.40          & 5.30           & 49.83          & 74.74          & 32.32          & 68.61 \\
        SimpleRL\textsuperscript{‡}         & 28.00          & 76.20          & 57.59          & 37.93          & 34.93          & 12.00          & 49.80          & 63.91          & 27.27          & 58.61 \\
        Oat\textsuperscript{‡}              & \textbf{36.00} & 78.40          & 59.75          & 42.52          & 36.40          & 10.00          & 52.92          & 59.89          & 33.84          & 56.13 \\
        LUFFY\textsuperscript{*}            & 21.30          & 77.80          & 44.82          & 40.00          & 36.40          & 14.67          & 50.69          & 81.83          & 32.32          & 74.67 \\
        LUFFY\textsuperscript{‡}            & 27.33          & \underline{83.20} & 60.24          & \underline{48.00} & 38.97          & \underline{17.33} &57.19& 81.83          & 35.86          & 75.19 \\
        
        \midrule
        
        \addlinespace[2pt]
        \textbf{\name\textsuperscript{*}}    & \underline{29.33} & 82.80          & \underline{61.69} & 47.41          & \textbf{43.38} & 17.30          & \underline{57.69} & \textbf{91.89} & \textbf{42.42} & \textbf{84.74} \\
        \textbf{\name\textsuperscript{‡}}    & \textbf{36.00} & \textbf{87.00} & \textbf{62.65} & \textbf{52.15} & 38.24          & \textbf{18.87} & \textbf{60.35} & 84.73          & 38.89          & 78.10 \\
        
        \bottomrule[1pt]
    \end{tabular}
    \end{adjustbox}
\end{table}

For the specialized Qwen-2.5-Math-7B model, as expected from its specialized design~\citep{yang2024qwen25mathtechnicalreportmathematical}, the Math model performs lower on the out-of-domain non-math benchmarks compared to the general-purpose Qwen-2.5-7B-Instruct. However, \name~not only leads the board in in-domain math tasks compared with baselines but also enables the Math model to achieve the highest out-of-domain test performance among all baselines. 
This suggests that the improvements are not solely due to domain-specific knowledge but may also reflect an enhancement of the model's general reasoning capabilities.

\subsection{Pass@k Evaluation}
% \citet{brown2024large} 
% We also utilize the pass@k metric to evaluate the model's reasoning capability~\citep{brown2024large}.
Many studies~\citep{chen2021evaluating,wang2022self} show that with a limited number of rollouts, models may perform poorly on certain tasks.
However, with a sufficiently large number of rollouts, they are more likely to sample good solutions for solving these problems. 
Therefore, to fully assess the model's potential performance, pass\@k accuracy (where $k$ is very large) is a more suitable metric~\citep{yue2025does}. 
In this context, a problem is considered solved if any of the $k$ sampled reasoning chains yields a correct answer.

\begin{figure}[t]
    \centering
    \begin{minipage}{0.45\textwidth}
        \centering
        \includegraphics[width=\textwidth]{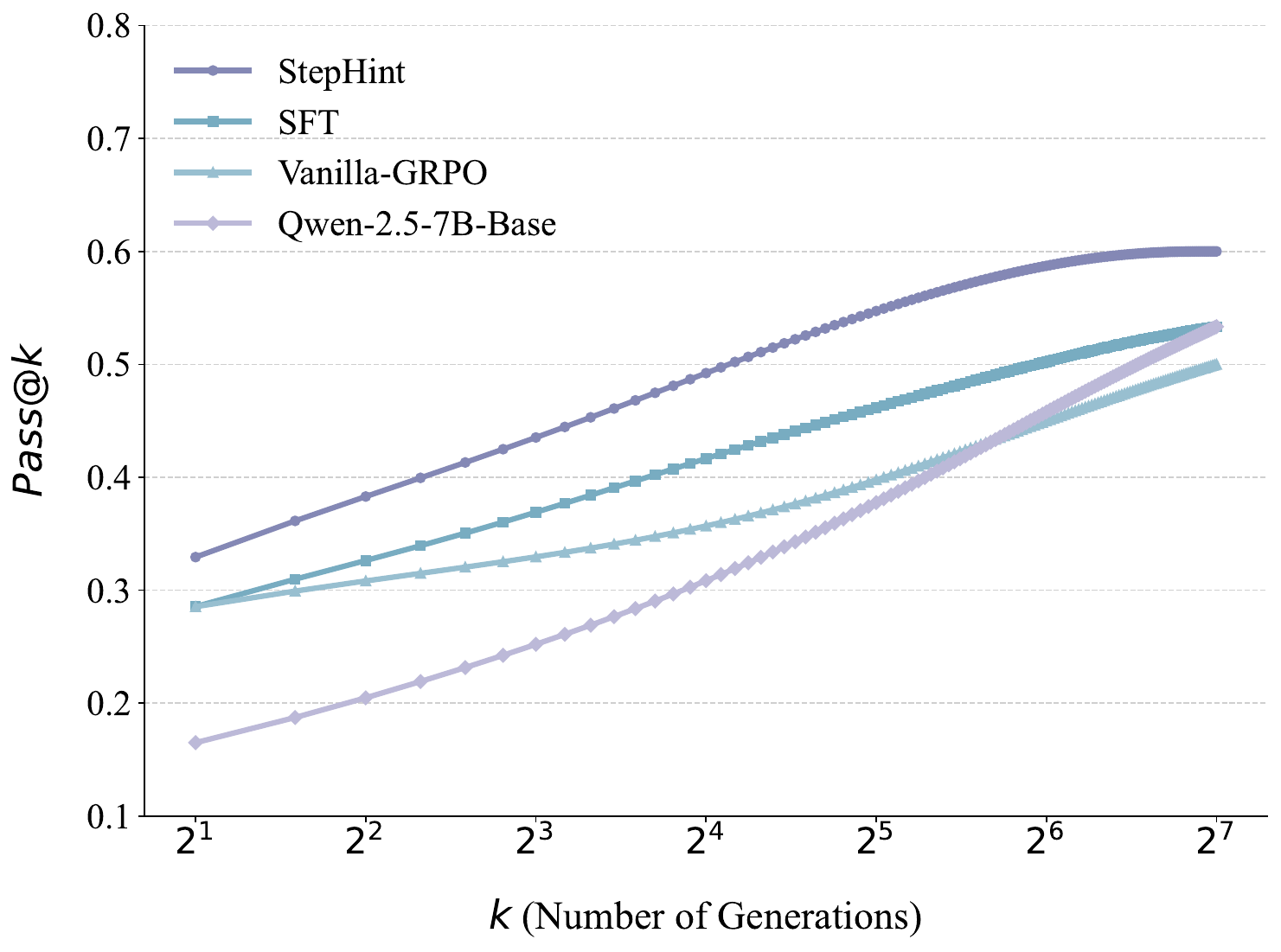}
        \captionsetup{labelformat=empty}
    \end{minipage}
    \hspace{6mm}
    \begin{minipage}{0.45\textwidth}
        \centering
        \includegraphics[width=\textwidth]{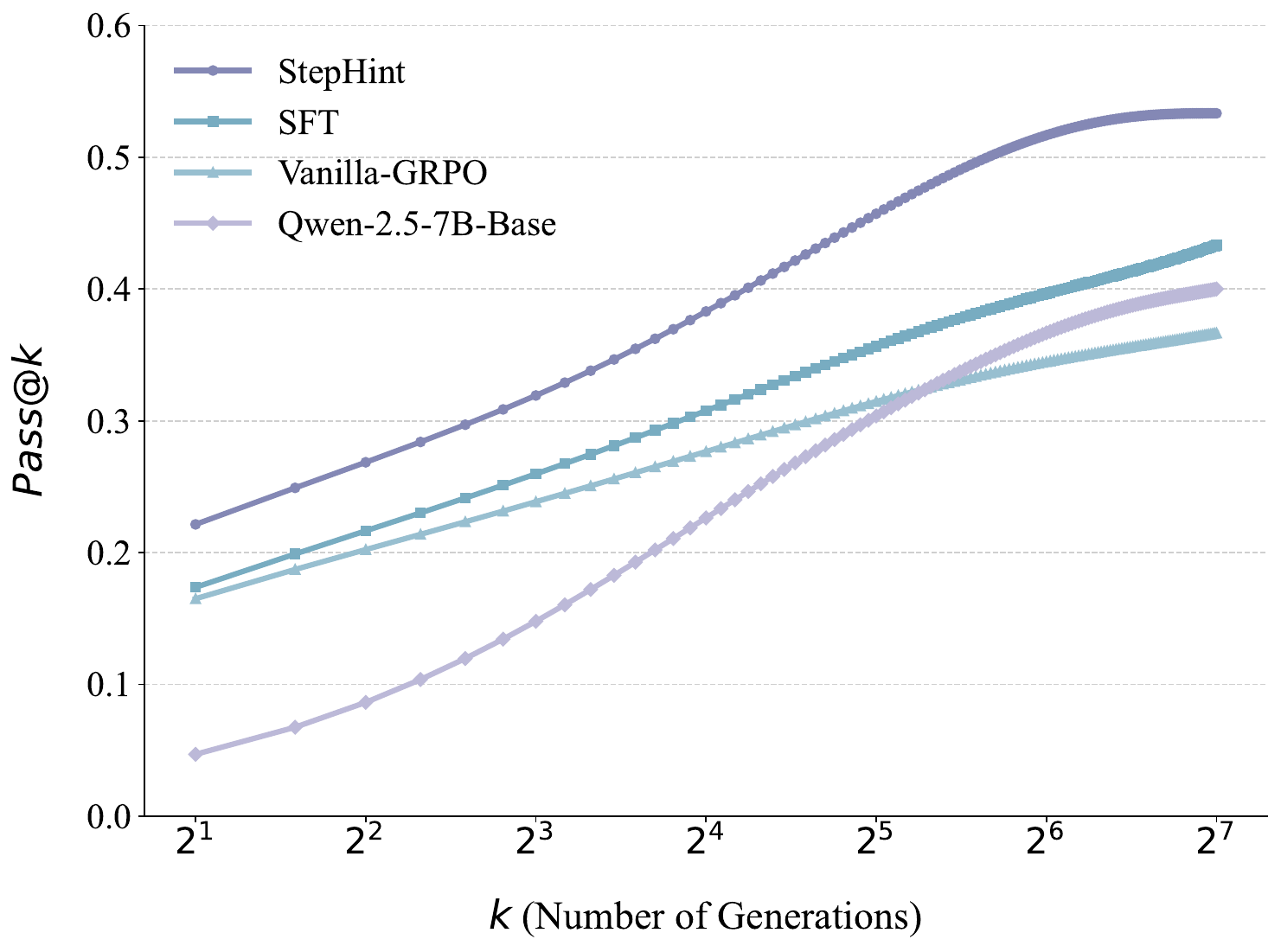}
        \captionsetup{labelformat=empty}
    \end{minipage}
    \caption{Comparison of pass@k results on the AIME24 and AIME25  datasets. \textbf{Left:} AIME24; \textbf{Right:} AIME25.}
    \label{fig:passk}
\end{figure}

Figure~\ref{fig:passk} presents the pass@k results on the AIME24 and AIME25 datasets. 
The results demonstrate that \name\ improves the model's pass@k performance as $k$ increases. 
In contrast, Vanilla-GRPO shows a slower rate of improvement at higher values of $k$, which aligns with findings from previous work~\citep{yue2025does}. 
The superior performance under pass@k evaluation further validates the effectiveness of \name.
Additionally, the performance difference can be attributed to the exploration strategies the models employ. 
While Vanilla-GRPO suffers from ``exploration stagnation,'' \name~guides the model's exploration, helping it break free from its ``comfort zone.''

\subsection{Method Analysis from Training Dynamics}

\begin{figure}[t]
    \centering
    \begin{minipage}{0.3\textwidth}
        \centering
        \includegraphics[width=\textwidth]{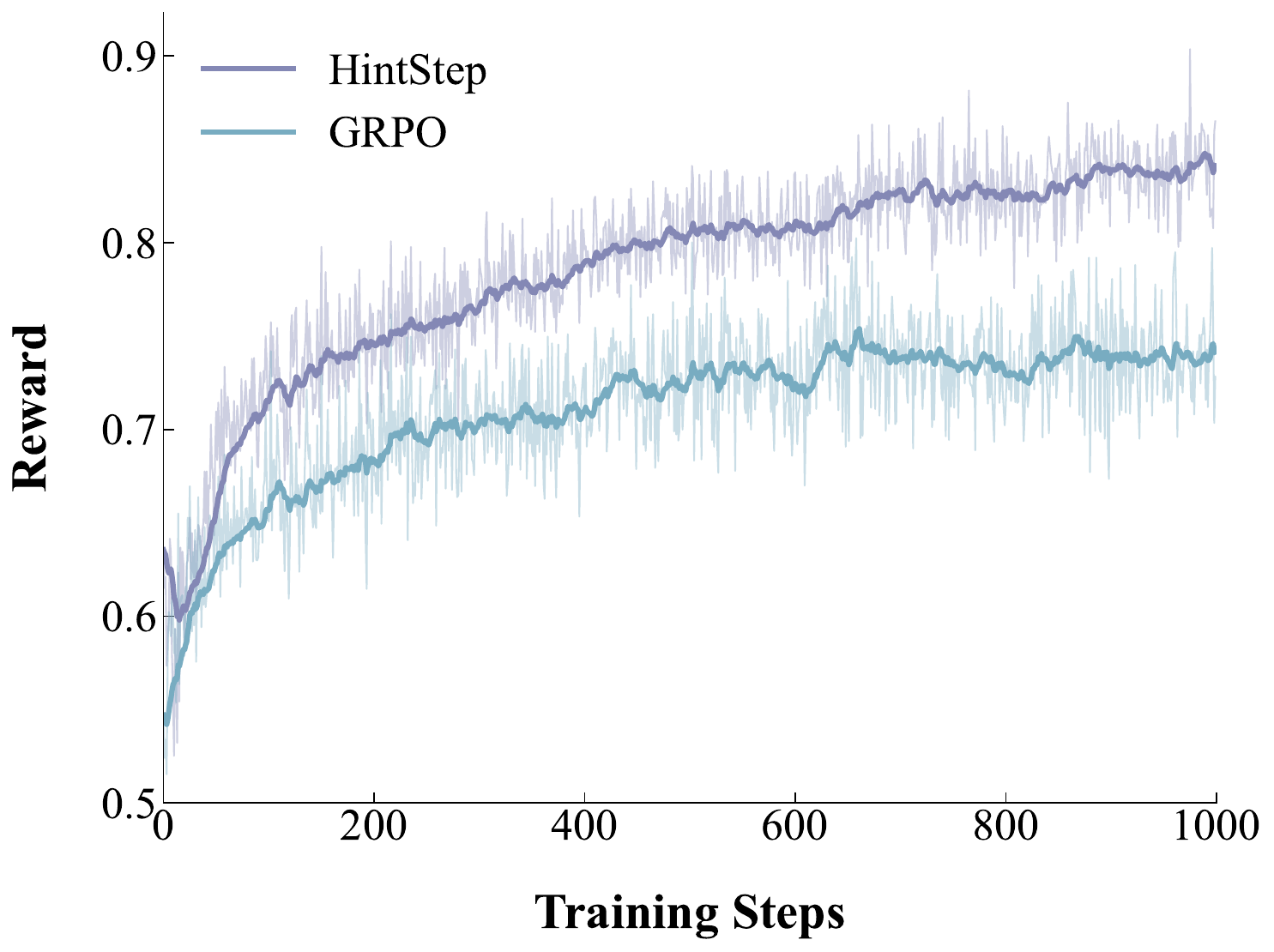}
        \captionsetup{labelformat=empty}
    \end{minipage}
    \hfill
    \begin{minipage}{0.3\textwidth}
        \centering
        \includegraphics[width=\textwidth]{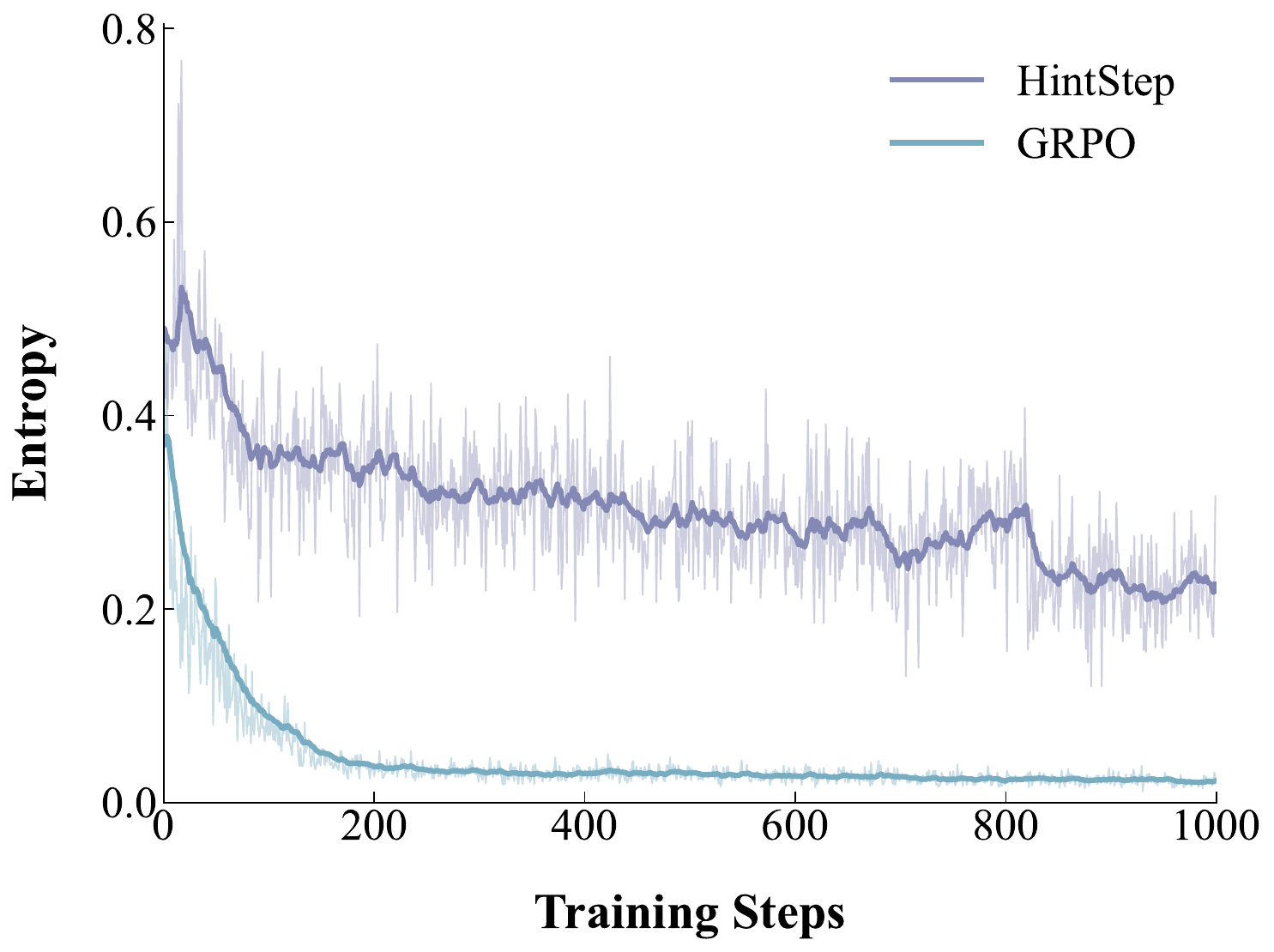}
        \captionsetup{labelformat=empty}
    \end{minipage}
    \hfill
    \begin{minipage}{0.3\textwidth}
        \centering
        \includegraphics[width=\textwidth]{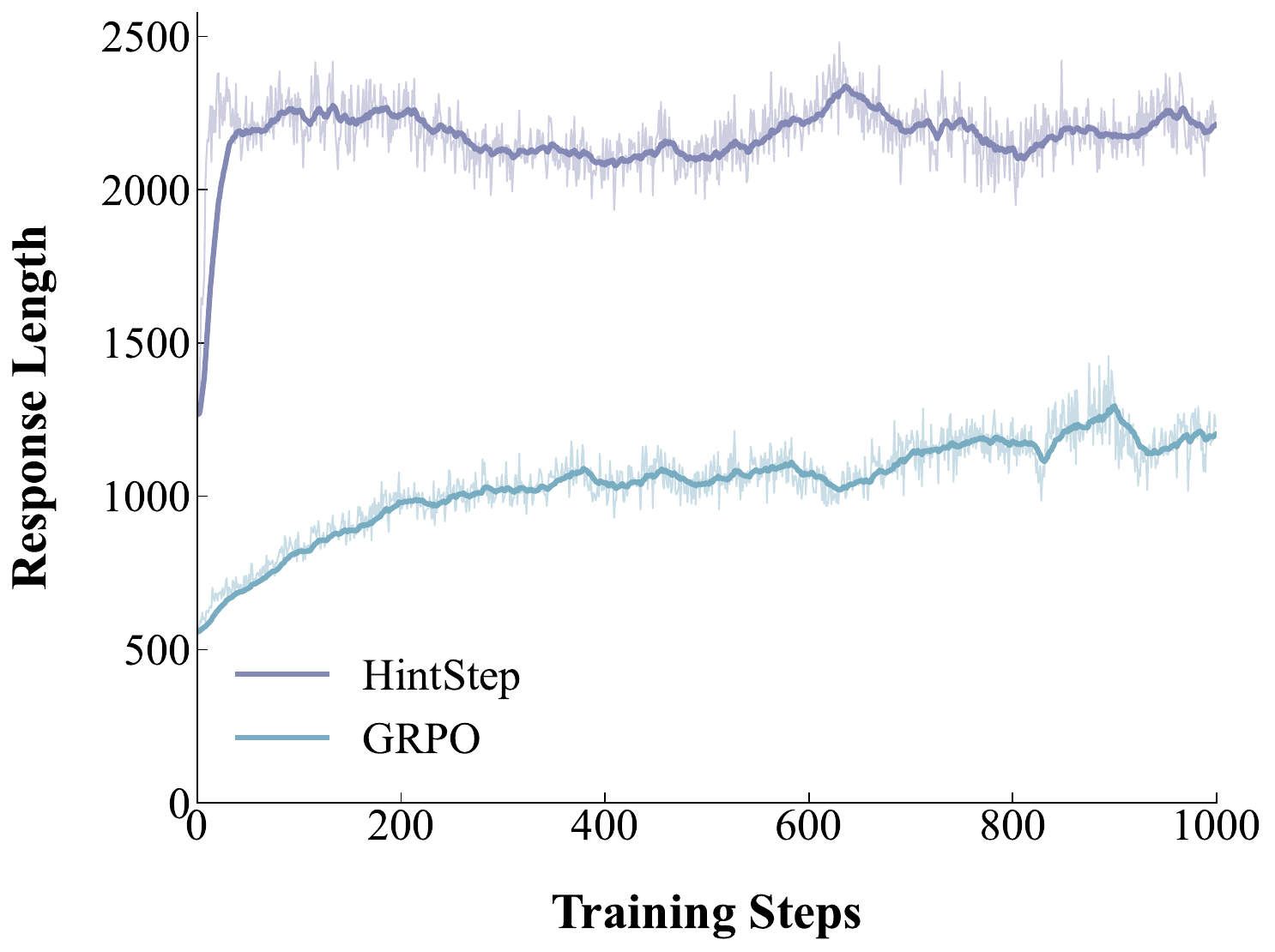}
        \captionsetup{labelformat=empty}
    \end{minipage}
    \caption{Training dynamics of \name~compared with GRPO. \textbf{Left:} Reward; \textbf{Middle:} Entropy; \textbf{Right:} Response Length.}
    \label{fig:traindy}
\end{figure}

We analyze the training dynamics by comparing \name~and Vanilla-GRPO on three key metrics: entropy, response length, and training rewards, as shown in Figure\ref{fig:traindy}.
The differences in these metrics offer valuable insights into the underlying behavior and effectiveness of \name.

\paragraph{Reward} The reward curves highlight the different learning phases. 
Due to the multi-level stepwise hints provided by \name, the problem-solving difficulty for the model is lower compared to Vanilla-GRPO. 
As a result, the reward score for \name\ is consistently higher, reflecting the mitigation of the near-miss reward issue.

A closer examination of the trends in both curves offers further insights.
Vanilla-GRPO shows a steady, monotonic increase in reward as it refines its existing policy. 
In contrast, \name\ experiences a brief initial dip in reward before a rapid and sustained increase. 
This initial dip likely reflects an adaptation period where the model transitions from simple exploitation to a more complex, hint-guided exploration.
Once adapted, the model efficiently discovers higher-reward solutions, leading to faster and effective learning to reason.

\paragraph{Entropy} Both methods exhibit an overall decrease in entropy, though their trajectories diverge as training progresses. 
The policy entropy for \name\ remains consistently higher than that of Vanilla-GRPO. 
This suggests that the hints provided by \name\ encourage a more diverse policy, preventing premature convergence to a narrow solution subspace and promoting a higher level of exploration~\citep{cheng2025reasoning}.
This trend reflects, to some extent, the mitigation of exploration stagnation.

\paragraph{Response Length} The two methods demonstrate distinct patterns in generated response length. 
\name~shows an initial sharp increase in length, which we attribute to the model learning to mimic the structured, stepwise reasoning chains provided by the multi-level hints. 
These hints are often more detailed than the model's initial, more direct responses.

These dynamics collectively illustrate that \name~fosters a more effective process for developing reasoning abilities.

% Collectively, these dynamics illustrate that \name~fosters a more effective learning process. Unlike the steady refinement of Vanilla-GRPO, \name~induces a structured exploration that, despite a brief initial adaptation, ultimately guides the model to better reasoning capabilities and higher performance.

\section{Related Works}

RL-based post-training has demonstrated remarkable success in mathematical reasoning tasks~\citep{shao2024deepseekmath, yang2024qwen25mathtechnicalreportmathematical}. Research in this area has primarily advanced in three directions: (1) optimizing the models and their training data, (2) refining inference-time strategies, and (3) improving policy optimization methods.

The first direction involves constructing high-quality, large-scale mathematical reasoning datasets~\citep{wang2023generative, ye2025limo} and designing specialized training or fine-tuning methods~\citep{jaech2024openai, mitra2024orca}. 
The second direction focuses on guiding the model's step-by-step thought processes without altering its underlying weights, typically through sophisticated prompting techniques such as Chain-of-Thought~\citep{wei2022chain} and innovations in in-context learning~\citep{wu2024beyond}. 
The third direction aims at developing advanced policy optimization algorithms. 
GRPO, an advancement of PPO~\citep{schulman2017proximal}, has recently gained popularity due to its simplicity and strong performance~\citep{hu2025open, zeng2025simplerl}. 
Additionally, several improvements have been proposed for GRPO; for example, \citet{liu2025understanding} identified inherent length and difficulty biases in vanilla GRPO and addressed these issues.

\citep{yan2025learning} is also related to this work. The authors use an \textit{entire} reasoning chain from stronger models as a reference trajectory. 
The reference is provided to the target model, which then independently generates rollouts several times without any hints. 
As a result, this approach does not fully address the near-miss reward issue, as the model still independently explores most of the time. 
While using reasoning chains from stronger models may help mitigate exploration stagnation, it inherently undermines the core exploration mechanism of the RL algorithm. 
In contrast, \name~better combines model-independent exploration with external hints, leading to more effective learning.

\section{Conclusion}
In this paper, we introduce \name, a novel RLVR algorithm that incorporates multi-level stepwise hints. 
This mechanism is designed to provide the model with assistance tailored to its evolving capabilities, thereby facilitating the learning process by addressing challenges such as near-miss rewards and exploration stagnation. 
\name~not only outperforms strong baselines on mathematical benchmarks but also demonstrates robust generalization to out-of-domain tasks, highlighting the promising potential of the stepwise hinting paradigm for RLVR enhancement.

%\newpage


\bibliography{iclr2025_conference}
\bibliographystyle{iclr2025_conference}

\appendix
\section{Appendix}

\subsection{Proof of Proposition~\ref{theo_chain}}
\label{proof1}
Proposition ~\ref{theo_chain}:
$$ \mathbb{E}_{t_k \sim p(\cdot|S_{k-1})} [H(\mathcal{R}|S_k)] \leq H(\mathcal{R}|S_{k-1}) $$
\begin{proof}
We want to prove the following inequality:
\[
\mathbb{E}_{t_k \sim p(\cdot|S_{k-1})} [H(\mathcal{R}|S_k)] \leq H(\mathcal{R}|S_{k-1})
\]
This inequality states that the expected entropy of solution space $\mathcal{R}$ conditioned on the state $S_k$ is less than or equal to the entropy of $\mathcal{R}$ conditioned on the prior state $S_{k-1}$. The state $S_k$ is reached from $S_{k-1}$ after an observation or transition $t_k$. Let's denote the random variable for this transition as $T_k$.

First, let's clarify the notation. The expression on the left-hand side, $\mathbb{E}_{t_k \sim p(\cdot|S_{k-1})} [H(\mathcal{R}|S_k)]$, represents the conditional entropy $H(\mathcal{R}|S_k)$. The conditional entropy $H(X|Y)$ is defined as the expectation of the entropy of $X$ over the values of $Y$. The subscript simply makes the underlying probability model explicit: the distribution of the state $S_k$ is induced by the distribution of the prior state $S_{k-1}$ and the transition $T_k$. 

The proof of this relationship relies on the non-negativity of conditional mutual information. The conditional mutual information between two random variables, $\mathcal{R}$ and $T_k$, given a third variable $S_{k-1}$, is defined as:
\[
I(\mathcal{R}; T_k | S_{k-1}) = H(\mathcal{R}|S_{k-1}) - H(\mathcal{R}|S_{k-1}, T_k)
\]
A fundamental property of mutual information is that it is always non-negative~\citep{mackay2003information,polyanskiy2025information}:
\[
I(\mathcal{R}; T_k | S_{k-1}) \ge 0
\]
From this, it directly follows that:
\[
H(\mathcal{R}|S_{k-1}) \ge H(\mathcal{R}|S_{k-1}, T_k)
\]
This equation shows that conditioning on an additional variable, $T_k$, can only decrease (or leave unchanged) the entropy of $\mathcal{R}$.

Now, we must relate the term $H(\mathcal{R}|S_{k-1}, T_k)$ to the term $H(\mathcal{R}|S_k)$. The state $S_k$ is the result of a process that starts in state $S_{k-1}$ and undergoes the transition $T_k$. This means that the information contained in the pair of variables $(S_{k-1}, T_k)$ fully determines the state $S_k$. In many typical models, knowing $S_k$ is equivalent to knowing the pair $(S_{k-1}, T_k)$ that produced it. If we assume this equivalence, then conditioning on $S_k$ is the same as conditioning on the pair $(S_{k-1}, T_k)$. Therefore, we have:
\[
H(\mathcal{R}|S_k) = H(\mathcal{R}|S_{k-1}, T_k)
\]

Substituting this equality back into our previous inequality, we arrive at the desired result:
\[
H(\mathcal{R}|S_k) \leq H(\mathcal{R}|S_{k-1})
\]
This completes the proof. 
\end{proof}

\subsection{Template}
\label{temp}
\begin{tcolorbox}[colframe=black, colback=gray!10, coltitle=black, boxrule=0.5mm, width=0.95\textwidth]
\textbf{Template} \\
\texttt{<|im\_start|>}system \\
You are a helpful assistant. The assistant first thinks about the reasoning process in the mind and then provides the user with the answer. The reasoning process and answer are enclosed within \texttt{<think> </think>} and \texttt{<answer> </answer>} tags, respectively, i.e., \texttt{<think>} reasoning process here \texttt{</think><answer>} answer here \texttt{</answer>}. Now the user asks you to solve a mathematical reasoning problem. After thinking, when you finally reach a solution, clearly state the answer marked with $\backslash\text{boxed}\{\}$ and within \texttt{<answer>} \texttt{</answer>} tags, i.e., \texttt{<answer>}$\backslash\text{boxed}\{\text{answer}\}$\texttt{</answer>}\\
\texttt{<|im\_end|>}\\
\texttt{<|im\_start|>} user\\
\textcolor{red}{\{\text{question}\}}\\
\texttt{<|im\_end|>}\\
\texttt{<|im\_start|>}assistant\\
\texttt{<think>}
\end{tcolorbox}

\subsection{Training Data Construction}
\label{dat}

For each question in the DAPO dataset~\citep{yu2025dapo}, we sample a total of 12 reasoning chains using DAPO-Qwen-32B~\citep{yu2025dapo}, QWQ-32B~\citep{qwq32b}, and DeepSeek-R1-Distill-Qwen-32B~\citep{guo2025deepseek}, with 4 samples per model. The sampling was conducted under a 0-shot setting, with a temperature of 1 and a maximum length of 8,192. We filter these to retain all reasoning chains that are both correct and have a length of no more than 4,110. In cases where multiple chains satisfy these conditions, we randomly select one.

\end{document}